\documentclass[conference]{IEEEtran}
\IEEEoverridecommandlockouts
\usepackage{cite}
\usepackage{amssymb,amsfonts}
\usepackage{tikz}
\usepackage{caption}
\usepackage{graphicx}
\usepackage{textcomp}
\usepackage{breqn}
\usepackage{amsthm}
\usepackage{algorithm}
\usepackage{algpseudocode}
\usepackage[hidelinks]{hyperref}
\usepackage{subcaption}

\newtheorem{theorem}{Theorem}
\newtheorem{lemma}{Lemma}

\newtheorem{assumption}{Assumption}
\theoremstyle{definition}
\newtheorem{definition}{Definition}
\theoremstyle{remark}

\definecolor{darkred}{RGB}{150,0,0}
\definecolor{darkgreen}{RGB}{0,150,0}
\definecolor{darkblue}{RGB}{0,0,150}
\hypersetup{colorlinks=true, linkcolor=red, citecolor=blue, urlcolor=darkblue}

\newcommand{\reals}{\mathbb{R}}
\newcommand{\calX}{\mathcal{X}}

\newcommand{\calN}{\mathcal{N}}

\begin{document}

\title{\LARGE \bf Multi-Agent Stage-wise Conservative Linear Bandits}

\author{Amirhossein Afsharrad$^1$, Ahmadreza Moradipari$^2$, Sanjay Lall$^1$
\thanks{$^1$ Stanford University, $^2$University of California, Santa Barbara. {\tt \small afsharrad@stanford.edu } }
}
\maketitle

\begin{abstract}
In many real-world applications such as recommendation systems, multiple learning agents must balance exploration and exploitation while maintaining safety guarantees to avoid catastrophic failures. We study the stochastic linear bandit problem in a multi-agent networked setting where agents must satisfy stage-wise conservative constraints. A network of $N$ agents collaboratively maximizes cumulative reward while ensuring that the expected reward at every round is no less than $(1-\alpha)$ times that of a baseline policy. Each agent observes local rewards with unknown parameters, but the network optimizes for the global parameter (average of local parameters). Agents communicate only with immediate neighbors, and each communication round incurs additional regret. We propose MA-SCLUCB (Multi-Agent Stage-wise Conservative Linear UCB), an episodic algorithm alternating between action selection and consensus-building phases. We prove that MA-SCLUCB achieves regret $\tilde{O}\left(\frac{d}{\sqrt{N}}\sqrt{T}\cdot\frac{\log(NT)}{\sqrt{\log(1/|\lambda_2|)}}\right)$ with high probability, where $d$ is the dimension, $T$ is the horizon, and $|\lambda_2|$ is the network's second largest eigenvalue magnitude. Our analysis shows: (i) collaboration yields $\frac{1}{\sqrt{N}}$ improvement despite local communication, (ii) communication overhead grows only logarithmically for well-connected networks, and (iii) stage-wise safety adds only lower-order regret. Thus, distributed learning with safety guarantees achieves near-optimal performance in reasonably connected networks.
\end{abstract}

\section{Introduction}

The stochastic linear bandit problem is a well-explored framework in sequential decision-making tasks that exhibit linear relationships, such as recommendation systems or path routing \cite{bubeck2012regret}. In this setting, an agent selects an action at each round and observes a random reward whose expected value depends linearly on the context of that action. The central objective is to maximize the cumulative reward obtained over $T$ time steps. In this work, we investigate the stage-wise constrained stochastic linear bandit problem in a multi-agent networked setting. Here, the network is also provided with a baseline policy that recommends an action at each stage, offering a guaranteed level of expected reward  \cite{vanroy,khezeli2019safe}. 

A group of $N$ agents aim to maximize their collective reward while ensuring that the expected reward of the chosen action at every round be no less than a fixed fraction of the expected reward from the given baseline policy. Each agent faces a local  linear bandit problem with unknown reward which may differ across agents, and needs to ensure performance at least as well as the baseline policy. The collective objective, however, is to identify the optimal action with respect to the global network parameters, defined as the averages of all individual reward and cost parameters. To reduce communication overhead, we impose two key assumptions: agents exchange information only with their immediate neighbors, and every communication step contributes additional regret. 

An example that might benefit from the design of stage-wise conservative learning algorithms arises in recommender systems, where the recommenders might wish to avoid   recommendations that are extremely disliked by the users at any single round.
Our proposed stage-wise conservative constraints ensure that at no round would the recommendation systems cause severe dissatisfaction for the user, and the reward of action employed by the learning algorithm, if not better, should be close to that of baseline policy.

\subsection{Previous work}

\textbf{Multi-armed Bandits.}
The multi-armed bandit (MAB) framework is a foundational model for sequential decision-making under uncertainty. It characterizes the exploration–exploitation dilemma, where a learner must balance selecting actions that yield high immediate rewards with exploring alternative actions to improve reward estimates over time \cite{bubeck2016multi}. Two widely used strategies have emerged for addressing this trade-off. The first is based on the optimism in the face of uncertainty (OFU) principle \cite{Auer, li2017provably, filippi2010parametric}, where Upper Confidence Bound (UCB) algorithms select the action–environment pair that appears optimal within the learner’s current confidence region. The second is Thompson Sampling (TS) \cite{thompson1933likelihood, kaufmann2012thompson, russo2016information, moradipari2018learning}, which maintains a posterior distribution over the unknown environment and randomly samples from it to determine the action to play.

\textbf{Linear Stochastic Bandits.} 
The study of linear stochastic bandits (LB) has led to a broad and well-established literature. Two widely used algorithms in this setting are Linear UCB (LUCB) and Linear Thompson Sampling (LTS). For LUCB, regret guarantees of order $\mathcal{O}(\sqrt{T} \log T)$ have been established \cite{dani2008stochastic, Tsitsiklis, abbasi2011improved}, while for LTS, bounds of order $\mathcal{O}(\sqrt{T} \log^{3/2} T)$ have been derived in the frequentist regime, where the unknown parameter is assumed to be fixed \cite{agrawal2013thompson, abeille2017linear}. Importantly, however, neither of these heuristics can be directly applied in our conservative setting.

\textbf{Conservativeness.}
The baseline model adopted in this paper was first proposed in \cite{vanroy,wu2016conservative} in the case of {\it cumulative constraints} on the reward. The stage-wise constraint was first studied in \cite{khezeli2019safe, moradipari2020stage}, where the learner's goal was to maximize cumulative reward while ensuring guaranteed level of the performance with respect to the given baseline policy at each step. In this work, we study a multi-agent setting: each agent faces a local linear bandit problem, but agents must collaborate to maximize the global network reward while simultaneously satisfying the stage-wise performance guarantee with respect to the baseline policy.

\textbf{Multi-agent Stochastic Bandits.}
Recent years have seen increasing attention on distributed and decentralized bandit problems. In the multi-armed bandit (MAB) setting, several studies have explored collaborative algorithms under communication or structural constraints. For example, UCB-based approaches such as coopUCB and coopUCB2 were proposed in \cite{DBLP:journals/corr/LandgrenSL16}, while \cite{9143736, ijcai2017p24} incorporated communication costs and decision trade-offs between pulling arms and sharing information. Other lines of work consider collisions, where multiple agents selecting the same arm receive reduced or no reward \cite{landgren2020distributed, 5738217}, or restrict play to a single agent per round with shared observations \cite{Kar2011BanditPI}. 
Our work differs from the aforementioned studies in that we consider a multi-agent setting where the agents’ goal is to maximize the global network reward, while their observations are limited to local parameters. Moreover, our setting is more restrictive, as agents must guarantee a certain level of performance at each step—i.e., no free exploration is allowed. 

\section{Preliminaries}\label{sec:prelim}

In this section, we present the notations and definitions used throughout the paper.

\textbf{Notations.} For a positive integer $n$, the set $\{1, 2, \ldots, n\}$ is denoted by $[n]$. For a vector $x \in \reals^d$ and positive definite matrix $\Sigma \in \reals^{d \times d}$, we define $\|x\|_{\Sigma} = \sqrt{x^T \Sigma x}$. The minimum eigenvalue of a matrix $A$ is denoted by $\lambda_{\min}(A)$. The identity matrix of dimension $d$ is denoted by $I_d$ or simply $I$ when the dimension is clear from context. The vector of all ones is denoted by $\mathbf{1}$.

\begin{definition}[Sub-Gaussian Random Variable]
\label{def:subgaussian}
A random variable $X$ with mean $\mathbb{E}[X] = \mu$ is said to be $R$-sub-Gaussian if for all $\lambda \in \reals$,
\begin{equation}
    \mathbb{E}[e^{\lambda (X - \mu)}] \leq \exp\left(\frac{\lambda^2 R^2}{2}\right).
\end{equation}
\end{definition}

\section{Problem Formulation}\label{sec:problem}

\subsection{Network Structure}

We consider a multi-agent network comprising $N$ agents operating over $T$ rounds. The network is represented as an undirected connected graph $G = (\mathcal{V}, \mathcal{E})$, where $\mathcal{V} = \{1, 2, \ldots, N\}$ is the set of agents (nodes) and $\mathcal{E}$ is the set of edges representing communication links. For each agent $i$, we denote by $\calN(i) = \{j : (i,j) \in \mathcal{E}\}$ the set of its neighbors.

The network structure is characterized by a doubly stochastic matrix $W \in \reals^{N \times N}$ where $W_{ij} \geq 0$ for all $i,j \in [N]$, and $W_{ij} = 0$ if and only if $j \notin \calN(i) \cup \{i\}$. The matrix $W$ satisfies the doubly stochastic property, meaning $\sum_{j=1}^N W_{ij} = 1$ and $\sum_{i=1}^N W_{ij} = 1$. The eigenvalues of $W$ satisfy $1 = \lambda_1 > |\lambda_2| \geq \cdots \geq |\lambda_N| \geq 0$. Each agent knows only its neighbors $\calN(i)$, the total number of agents $N$, and the second largest eigenvalue in absolute value $|\lambda_2|$.

\subsection{Local Bandit Problems}

Each agent $i \in [N]$ has its own local linear bandit problem characterized by an unknown reward parameter $\theta^i_* \in \reals^d$. At each round $t \in [T]$, when an action $x_t \in \calX$ is played by the network, each agent $i$ observes a local reward
\begin{equation}
    r^i_t = x_t^\top \theta^i_* + \eta^i_t,
\end{equation}
where $\calX \subset \reals^d$ is a convex and compact action set available to all agents, and $\eta^i_t$ is the observation noise for agent $i$ at time $t$.

\subsection{Global Objective and Conservative Constraints}

The global reward parameter is defined as the average of all local parameters
\begin{equation}
    \theta_*^{\text{global}} = \frac{1}{N}\sum_{i=1}^N \theta_*^i.
\end{equation}

The network is provided with a baseline policy that suggests actions $x_{b,t} \in \calX$ at each round $t$. The expected reward of the baseline action with respect to the global parameter is
\begin{equation}
    r_{b,t} = x_{b,t}^\top \theta_*^{\text{global}}.
\end{equation}
We assume that the values $r_{b,t}$ are known to all agents, for example from historical data.

\textbf{Stage-wise Conservative Constraint.} At each round $t$, the action $x_t$ chosen by the network must satisfy
\begin{equation}\label{eq:constraint}
    x_t^\top \theta_*^{\text{global}} \geq (1-\alpha) r_{b,t},
\end{equation}
where $\alpha \in (0,1)$ is the conservatism parameter. An action satisfying \eqref{eq:constraint} is called \emph{safe}. Note that the positivity of $r_{b,t}$ is guaranteed by Assumption~\ref{ass:baseline} in Section~\ref{subsec:assumption}.

\subsection{Action Selection Protocol}

At each round $t$, the network coordinator randomly selects an agent index $a(t) \in [N]$ uniformly at random. Agent $a(t)$ then selects an action $x_t \in \calX$ based on its current knowledge. All agents play the action $x_t$ simultaneously, and each agent $i$ observes its local reward $r^i_t$.

\subsection{Objective}

The goal is to minimize the cumulative pseudo-regret while satisfying the conservative constraint \eqref{eq:constraint}. The cumulative pseudo-regret is defined as
\begin{equation}
    \mathcal{R}(T) = \sum_{t=1}^T \left[x^{*\top} \theta_*^{\text{global}} - x_t^\top \theta_*^{\text{global}}\right],
\end{equation}
where $x^* = \arg\max_{x \in \calX} x^\top \theta_*^{\text{global}}$ is the optimal action.

\subsection{Assumptions}
\label{subsec:assumption}

\begin{assumption}[Sub-Gaussian Noise]\label{ass:noise}
For all $t \in [T]$ and $i \in [N]$, the noise variables $\eta^i_t$ are conditionally zero-mean and $R$-sub-Gaussian given the filtration $\mathcal{F}_{t-1}$ containing all information up to round $t-1$. That is, $\mathbb{E}[\eta^i_t | \mathcal{F}_{t-1}] = 0$ and $\eta^i_t | \mathcal{F}_{t-1}$ is $R$-sub-Gaussian in the sense of Definition~\ref{def:subgaussian}.
\end{assumption}

\begin{assumption}[Bounded Parameters]\label{ass:param}
Let $S$ denote an upper bound on the norm of the local parameters, that is, $\|\theta^i_*\|_2 \leq S$ for all $i \in [N]$. We assume that $S$ is independent of the number of agents $N$, meaning that even as the network size grows, the individual parameter norms remain uniformly bounded by the same constant $S$.
\end{assumption}

\begin{assumption}[Bounded Actions]\label{ass:action}
The action set $\calX$ is compact and convex. Due to compactness, $L > 0$ exists such that $\|x\|_2 \leq L$ for all $x \in \calX$. Moreover, we assume that $x^\top \theta_*^{\text{global}} \in [0,1]$ for all $x \in \calX$.
\end{assumption}

\begin{assumption}[Baseline Bounds]\label{ass:baseline}
Let $\kappa_{b,t} = x^{*\top} \theta_*^{\text{global}} - r_{b,t}$ denote the sub-optimality gap of the baseline action at time $t$. We assume there exist constants $\kappa_l$, $\kappa_h$, $r_l$, and $r_h$ such that $0 \leq \kappa_l \leq \kappa_{b,t} \leq \kappa_h$ and $0 < r_l \leq r_{b,t} \leq r_h$ for all $t \in [T]$. These bounds are independent of the horizon $T$, ensuring that the baseline policy maintains consistent quality regardless of the time horizon.
\end{assumption}

\section{Algorithm Description}\label{sec:algorithm}

We present the Multi-Agent Stage-wise Conservative Linear UCB (MA-SCLUCB) algorithm. To balance exploration, exploitation, and communication needs, MA-SCLUCB operates in an episodic structure. During each episode, the network first selects and plays an action, then engages in communication among agents to share information about the observed rewards.

\subsection{Episode Structure}

The algorithm divides time into episodes indexed by $s = 1, 2, \ldots$. Each episode $s$ begins at time $t_s$ and consists of two phases. In the exploration-exploitation phase, the network selects and plays a single action $x_{t_s}$ based on current knowledge. All agents observe their local rewards from this action. In the subsequent communication phase, agents exchange information with their neighbors over $q(s)$ rounds to compute estimates of the average reward across the network. During communication, all agents continue playing the same action $x_{t_s}$ to maintain consistency, though this incurs additional regret. The length of the communication phase $q(s)$ grows logarithmically with the episode number to ensure increasingly accurate consensus as the algorithm progresses. Specifically, we set
\begin{equation}\label{eq:comm_length}
    q(s) = \left\lceil \frac{\log(2Ns)}{\sqrt{2\log(1/|\lambda_2|)}} \right\rceil,
\end{equation}
where $|\lambda_2|$ is the second largest eigenvalue of the network's weight matrix $W$ in absolute value. This schedule guarantees that consensus errors decay at an appropriate rate, which is crucial for maintaining valid confidence regions.

\subsection{Information Flow and Estimation}

After $s$ episodes, each agent $i$ maintains estimates of the global parameter based on the history of played actions and observed rewards. Let $x_{t_1}, \ldots, x_{t_s}$ denote the actions played at the start of each episode.

In episode $s$, each agent $j$ initially observes its local reward $r^j_{t_s} = x_{t_s}^\top \theta^j_* + \eta^j_{t_s}$. During the communication phase, agents apply the accelerated consensus protocol (Algorithm~\ref{alg:mix}) to these local observations. After $q(s)$ communication rounds, agent $i$ obtains an estimate $y^i_s$ that approximates the average reward $\frac{1}{N}\sum_{j=1}^N r^j_{t_s}$ with an approximation error of order $1/s$. The precise characterization of this approximation error and its impact on the confidence regions will be analyzed in Section~\ref{sec:regret}.

\textbf{Regularized Least Squares Estimation.} Using the history of actions and reward estimates, each agent $i$ maintains a regularized least squares estimate of the global parameter. The Gram matrix after $s$ episodes is
\begin{equation}
    \Sigma_{s} = \lambda I + \sum_{k=1}^{s} x_{t_k} x_{t_k}^\top,
\end{equation}
where $\lambda > 0$ is the regularization parameter. Agent $i$ then computes its estimate as
\begin{equation}\label{eq:rls}
    \hat{\theta}^{\text{global},i}_{s} = \Sigma_{s}^{-1} \sum_{k=1}^{s} x_{t_k} y^i_k.
\end{equation}

\textbf{Confidence Regions.} Each agent $i$ constructs a confidence ellipsoid around its estimate to account for estimation uncertainty
\begin{equation}\label{eq:confidence}
    \mathcal{E}^i_{s} = \left\{ \theta \in \reals^d : \|\theta - \hat{\theta}^{\text{global},i}_{s}\|_{\Sigma_s} \leq \beta_s \right\},
\end{equation}
where the confidence radius accounts for both observation noise and consensus errors
\begin{equation}\label{eq:beta}
    \beta_s = \frac{R}{\sqrt{N}}\sqrt{d\log\left(\frac{1 + sL^2/\lambda}{\delta}\right)} + \sqrt{\lambda}S + \frac{L}{\sqrt{\lambda}}.
\end{equation}
The $\frac{R}{\sqrt{N}}$ term reflects the variance reduction from averaging $N$ agents' observations, while the $\frac{L}{\sqrt{\lambda}}$ term accounts for consensus approximation errors.

\subsection{Action Selection}

At the start of episode $s$, the network must select an action that balances exploration and exploitation while ensuring safety.

\textbf{Estimated Safe Set.} Given its confidence region, agent $i$ constructs the estimated safe set as the set of actions guaranteed to satisfy the conservative constraint for all parameters in the confidence region
\begin{equation}\label{eq:safe_set}
    \mathcal{X}^{\text{safe},i}_{s} = \left\{ x \in \calX : \min_{v \in \mathcal{E}^i_{s}} x^\top v \geq (1-\alpha) r_{b,t_s} \right\}.
\end{equation}
Using the ellipsoid structure, this simplifies to
\begin{equation}
    \mathcal{X}^{\text{safe},i}_{s} = \left\{ x \in \calX : x^\top \hat{\theta}^{\text{global},i}_{s} - \beta_s \|x\|_{\Sigma_s^{-1}} \geq (1-\alpha) r_{b,t_s} \right\}.
\end{equation}

\textbf{UCB Action Selection.} When agent $a(s)$ is selected to choose the action for episode $s$, it first checks whether sufficient exploration has occurred. If the minimum eigenvalue satisfies $\lambda_{\min}(\Sigma_{s-1}) \geq k_{t_s}$ where
\begin{equation}
    k_{t_s} = \left(\frac{2L\beta_{s-1}}{\kappa_l + \alpha r_l}\right)^2,
\end{equation}
and the safe set is non-empty, the agent selects the optimistic action within the safe set by solving
\begin{equation}
    \max_{x \in \mathcal{X}^{\text{safe},a(s)}_{s-1}} \left[ x^\top \hat{\theta}^{\text{global},a(s)}_{s-1} + \beta_{s-1} \|x\|_{\Sigma_{s-1}^{-1}} \right].
\end{equation}
This optimization problem is convex since the objective is convex as the sum of linear and convex functions, the safe set constraint is convex as it is defined by a linear inequality, and the action set $\calX$ is convex by assumption.

\textbf{Conservative Action Construction.} When insufficient exploration has occurred or the safe set is empty, the algorithm plays a conservative action that guarantees safety while promoting exploration
\begin{equation}\label{eq:conservative}
    x^{\text{cons}}_{t} = (1-\rho) x_{b,t} + \rho \zeta_{t},
\end{equation}
where $\zeta_t$ is a random exploration vector sampled uniformly from the unit sphere, and $\rho = \frac{\alpha r_l}{S + r_h}$ is chosen to ensure safety. The random component ensures that the covariance satisfies $\lambda_{\min}(\text{Cov}(\zeta_t)) = \sigma_{\zeta}^2 > 0$, promoting exploration in all directions.

\subsection{Communication Protocol}

During the communication phase, agents use an accelerated consensus protocol to efficiently estimate average rewards. The protocol leverages the spectral properties of the network to achieve consensus with minimal communication rounds.

\begin{algorithm}
\caption{Accelerated Consensus Mix Function}\label{alg:mix}
\begin{algorithmic}[1]
\Function{Mix}{$\alpha_h^i, h, i, [W_{ij}]_{j=1}^N, |\lambda_2|$}
    \If{$h = 0$}
        \State $c_0 \gets 1/2$, $c_{-1} \gets 0$
        \State $\alpha_0^i \gets \alpha_0^i / 2$
        \State $\alpha_{-1}^i \gets 0$ \Comment{Initialize to zero vector}
    \EndIf
    \State Send $\alpha_h^i$ to all neighbors $j \in \calN(i)$
    \State Receive $\alpha_h^j$ from all neighbors $j \in \calN(i)$
    \State $z_h^i \gets \sum_{j \in \calN(i) \cup \{i\}} 2W_{ij} \alpha_h^j / |\lambda_2|$
    \State $c_{h+1} \gets 2c_h / |\lambda_2| - c_{h-1}$
    \State $\alpha_{h+1}^i \gets \frac{c_h}{c_{h+1}} z_h^i - \frac{c_{h-1}}{c_{h+1}} \alpha_{h-1}^i$
    \If{$h = 0$}
        \State $c_0 \gets 2c_0$, $\alpha_0^i \gets 2\alpha_0^i$
    \EndIf
    \State \Return $\alpha_{h+1}^i$
\EndFunction
\end{algorithmic}
\end{algorithm}

The consensus protocol works as follows. Each agent $i$ initializes a value $v_0^i$ with its local reward $r^i_{t_s}$. Then, for $h = 0$ to $q(s) - 1$, agent $i$ iteratively updates its value by calling $v_{h+1}^i \gets \text{Mix}(v_h^i, h, i, [W_{ij}]_{j=1}^N, |\lambda_2|)$ as shown in Algorithm~\ref{alg:mix}. During these iterations, all agents continue playing the same action $x_{t_s}$. After $q(s)$ iterations, each agent $i$ obtains its final estimate $y^i_s = v_{q(s)}^i$, which approximates the network average $\frac{1}{N}\sum_{j=1}^N r^j_{t_s}$.

The complete MA-SCLUCB algorithm is presented in Algorithm~\ref{alg:main}.

\begin{algorithm}
\caption{MA-SCLUCB: Multi-Agent Stage-wise Conservative Linear UCB}\label{alg:main}
\begin{algorithmic}[1]
\Require $\delta, T, \lambda, \alpha, N, |\lambda_2|, W$
\State Initialize: $t \gets 1$, $s \gets 1$
\While{$t < T$}
    \State \textbf{Episode $s$ begins:}
    \State $t_s \gets t$ \Comment{Start time of episode $s$}
    \State $q(s) \gets \lceil \log(2Ns) / \sqrt{2\log(1/|\lambda_2|)} \rceil$
    
    \State \textbf{Exploration-Exploitation Phase:}
    \State Network coordinator selects agent $a(s)$ uniformly at random
    \State Agent $a(s)$ computes $\hat{\theta}^{\text{global},a(s)}_{s-1}$ using \eqref{eq:rls}
    \State Agent $a(s)$ constructs confidence region $\mathcal{E}^{a(s)}_{s-1}$ using \eqref{eq:confidence}
    \State Agent $a(s)$ computes safe set $\mathcal{X}^{\text{safe},a(s)}_{s-1}$ using \eqref{eq:safe_set}
    
    \State $k_{t_s} \gets \left(\frac{2L\beta_{s-1}}{\kappa_l + \alpha r_l}\right)^2$
    
    \If{$\mathcal{X}^{\text{safe},a(s)}_{s-1} \neq \emptyset$ \textbf{and} $\lambda_{\min}(\Sigma_{s-1}) \geq k_{t_s}$}
        \State \textbf{UCB Action Selection:}
        \State $(\bar{x}_{t_s}, \bar{\theta}_{t_s}) \gets \arg\max\limits_{\substack{x \in \mathcal{X}^{\text{safe},a(s)}_{s-1} \\ \theta \in \mathcal{E}^{a(s)}_{s-1}}} x^\top \theta$
        \State $x_{t_s} \gets \bar{x}_{t_s}$
    \Else
        \State \textbf{Conservative Action:}
        \State Sample $\zeta_{t_s}$ uniformly from unit sphere
        \State $x_{t_s} \gets (1-\rho) x_{b,t_s} + \rho \zeta_{t_s}$ where $\rho = \frac{\alpha r_l}{S + r_h}$
    \EndIf
    
    \State All agents play $x_{t_s}$
    \State Each agent $i$ observes $r^i_{t_s} = x_{t_s}^\top \theta^i_* + \eta^i_{t_s}$
    
    \If{$t_s + q(s) > T$}
        \State \textbf{break} \Comment{Not enough time for communication}
    \EndIf
    
    \State \textbf{Communication Phase:}
    \For{each agent $i \in [N]$ in parallel}
        \State $v_0^i \gets r^i_{t_s}$ \Comment{Initialize with local reward}
    \EndFor
    
    \For{$h = 0$ to $q(s) - 1$}
        \For{each agent $i \in [N]$ in parallel}
            \State $v_{h+1}^i \gets \text{Mix}(v_h^i, h, i, [W_{ij}]_{j=1}^N, |\lambda_2|)$
        \EndFor
        \State All agents play $x_{t_s}$ \Comment{Same action during communication}
        \State $t \gets t + 1$
    \EndFor
    
    \For{each agent $i \in [N]$}
        \State $y^i_s \gets v_{q(s)}^i$ \Comment{Final consensus estimate}
        \State Update: $\Sigma_s \gets \lambda I + \sum_{k=1}^s x_{t_k} x_{t_k}^\top$
        \State Update: $\hat{\theta}^{\text{global},i}_s \gets \Sigma_s^{-1} \sum_{k=1}^s x_{t_k} y^i_k$
    \EndFor
    
    \State $s \gets s + 1$
\EndWhile
\end{algorithmic}
\end{algorithm}

\section{Regret Analysis}\label{sec:regret}

We analyze MA-SCLUCB in two steps. First, we establish that the accelerated consensus protocol provides accurate estimates and prove that the optimal action belongs to every agent's estimated safe set once sufficient exploration has occurred. Second, we derive the overall regret bound by decomposing episodes into UCB and conservative episodes and accounting for the communication overhead.

Throughout, episodes are indexed by $s=1,2,\dots$ with start times $(t_s)_s$, and $M$ denotes the number of episodes completed by time $T$. We recall $\Sigma_s$, $\hat{\theta}^{\text{global},i}_s$, $\mathcal{E}^i_s$, $\mathcal{X}^{\text{safe},i}_s$, $\beta_s$ and $q(s)$ from \eqref{eq:rls}–\eqref{eq:comm_length}.

\subsection{Main Results}\label{subsec:regret-main}

We first establish the accuracy of the consensus protocol.

\begin{lemma}[Consensus Accuracy]\label{lem:consensus}
Let $W$ be the doubly stochastic weight matrix with second largest eigenvalue (in absolute value) $|\lambda_2| < 1$. After $q(s) = \lceil \log(2Ns) / \sqrt{2\log(1/|\lambda_2|)} \rceil$ communication rounds using the accelerated consensus protocol (Algorithm~\ref{alg:mix}), each agent $i$ obtains an estimate $y^i_s$ satisfying
\begin{equation}
    \left|y^i_s - \frac{1}{N}\sum_{j=1}^N r^j_{t_s}\right| \leq \frac{1}{s}.
\end{equation}
\end{lemma}

\begin{theorem}[Distributed Confidence Sets and Safe-Set Inclusion]\label{thm:confidence}
Fix $\delta\in(0,1)$ and set $\delta_{\mathrm{conf}}:=\delta/(2N)$. Use the communication schedule $q(s)$ in \eqref{eq:comm_length} and compute $\beta_s$ as in \eqref{eq:beta} with $\delta$ replaced by $\delta_{\mathrm{conf}}$. Then, with probability at least $1-\delta/2$, the following hold simultaneously for all $s\le M$ and $i\in[N]$:
\begin{enumerate}
\item[\emph{(i)}] \textbf{Valid confidence sets (distributed).} $\theta_*^{\mathrm{global}}\in \mathcal{E}^i_s$.
\item[\emph{(ii)}] \textbf{Safe-set inclusion once excited.} If
\begin{equation}\label{eq:lambda-thresh}
\lambda_{\min}(\Sigma_{s-1})\ \ge\ \Big(\tfrac{2L\,\beta_{s-1}}{\kappa_l+\alpha r_l}\Big)^{\!2},
\end{equation}
then $x^\star\in\mathcal{X}^{\mathrm{safe},i}_{s-1}$ for every agent $i$. Consequently, the UCB optimizer
\[
(\bar x_{t_s},\bar\theta_{t_s})\in
\arg\max_{x\in\mathcal{X}^{\mathrm{safe},i}_{s-1},\ \theta\in\mathcal{E}^i_{s-1}}
x^\top\theta
\]
is optimistic within the safe set: $\bar x_{t_s}^\top\bar\theta_{t_s}\ge x^{*\top}\theta_*^{\mathrm{global}}$.
\end{enumerate}
\end{theorem}

\begin{lemma}[Number of Conservative Episodes]\label{lem:num-cons}
Let $N_M^c:=|\{s\le M:\ \text{episode $s$ plays the conservative action \eqref{eq:conservative}}\}|$. With $\rho=\alpha r_l/(S+r_h)$ and $h_1=2\rho(1-\rho)L+2\rho^2$, we have, with probability at least $1-\delta/2$,
\begin{align}\label{eq:num-cons}
N_M^c\ \le\
&\ \Big(\tfrac{2L\,\beta_M}{\rho\,\sigma_\zeta(\kappa_l+\alpha r_l)}\Big)^{\!2}
\ +\ \frac{2h_1^2}{\rho^4\sigma_\zeta^4}\log\tfrac{d}{\delta/2} \\[-2pt]
&\ +\ \frac{2Lh_1\,\beta_M}{\rho^3\sigma_\zeta^3(\kappa_l+\alpha r_l)}
   \sqrt{\,8\log\tfrac{d}{\delta/2}\,}. \nonumber
\end{align}
\end{lemma}

\begin{theorem}[High-Probability Regret of MA-SCLUCB]\label{thm:regret}
Run MA-SCLUCB with $q(s)$ from \eqref{eq:comm_length}. Let $M$ be the number of episodes by time $T$. With probability at least $1-\delta$,
\begin{align}\label{eq:regret-main}
\mathcal{R}(T)\ \le\ (1+q(M))\Big[
&\ 2\beta_M\sqrt{\,2dM\log\!\big(1+\tfrac{ML^2}{\lambda d}\big)}\\[-2pt]
&\ +\ N_M^c\cdot\big(\kappa_h+\rho(r_h+S)\big)\Big],\nonumber
\end{align}
where $N_M^c$ satisfies \eqref{eq:num-cons}.
\end{theorem}

\textbf{Orders of Magnitude and Intuition.}
Since each episode requires at least $1+q(1)$ rounds, we have $M \le T/(1+q(1))$. Using this bound and noting that $q(M) \le q(T/(1+q(1))) = O(\log(NT)/\sqrt{\log(1/|\lambda_2|)})$, the regret bound simplifies to:
\[
\mathcal{R}(T) = \tilde{O}\!\left(
\frac{d}{\sqrt{N}}\sqrt{T}\,
\Big(1+\frac{\log(NT)}{\sqrt{\log(1/|\lambda_2|)}}\Big)\right).
\]

This bound reveals three key insights about multi-agent learning with safety constraints:

\textbf{Network advantage.} The $\frac{1}{\sqrt{N}}$ factor shows that collaboration provides a fundamental statistical advantage. Each agent observes noisy rewards with variance $R^2$, but averaging across $N$ agents reduces the effective variance to $R^2/N$. This directly translates to the $\frac{d}{\sqrt{N}}\sqrt{T}$ term, improving over the single-agent bound of $d\sqrt{T}$. Notably, this improvement occurs despite agents only communicating locally.

\textbf{Communication price.} The factor $(1+\frac{\log(NT)}{\sqrt{\log(1/|\lambda_2|)}})$ captures the cost of achieving consensus through local communication, where the "1" represents the baseline cost of learning and the logarithmic term represents the additional communication overhead. Better connected networks (smaller $|\lambda_2|$) require fewer consensus rounds, reducing this overhead. For well-connected networks where $|\lambda_2|$ is bounded away from 1, this additive communication term grows only logarithmically in $NT$, making it a small price to pay for the $\sqrt{N}$ improvement from collaboration.

\textbf{Safety is cheap.} The conservative episodes contribute only $\tilde{O}(d\log T/N)$ to the regret—a lower-order term. This shows that maintaining stage-wise safety does not fundamentally change the regret scaling; the algorithm can guarantee safety at every step while preserving the $\sqrt{T}$ growth rate. The $1/N$ factor here further demonstrates that larger networks better amortize the exploration cost of ensuring safety.

Together, these factors show that distributed learning with safety constraints achieves near-optimal scaling when the network is reasonably connected, with the multi-agent collaboration more than compensating for the communication overhead.

\subsection{Proofs}\label{subsec:regret-proofs}

\textbf{Proof of Lemma~\ref{lem:consensus}.}
The accelerated consensus protocol uses polynomial approximation to suppress non-consensus eigenmodes of $W$. After $q(s)$ iterations, the polynomial $p_{q(s)}(W)$ satisfies $\|p_{q(s)}(W) - \frac{1}{N}\mathbf{1}\mathbf{1}^\top\|_2 \le 1/(Ns)$ by construction of the schedule. For any initial vector of rewards $r_s = [r^i_{t_s}]_{i=1}^N$ with $\|r_s\|_2 \le N$ (which holds under Assumption~\ref{ass:action} since each component is bounded by 1), we have 
\begin{align*}
\|p_{q(s)}(W)r_s - \tfrac{1}{N}\mathbf{1}\mathbf{1}^\top r_s\|_2 &\le \|p_{q(s)}(W) - \tfrac{1}{N}\mathbf{1}\mathbf{1}^\top\|_2 \cdot \|r_s\|_2 \\
&\le \tfrac{1}{Ns} \cdot N = \tfrac{1}{s}.
\end{align*}

Since $y^i_s$ is the $i$-th component of $p_{q(s)}(W)r_s$ and $\frac{1}{N}\mathbf{1}^\top r_s$ is the true average, each component satisfies the bound $|y^i_s - \frac{1}{N}\sum_{j=1}^N r^j_{t_s}| \le 1/s$.

\textbf{Proof of Theorem~\ref{thm:confidence}.}
\emph{(i) Distributed confidence sets.}
From Lemma~\ref{lem:consensus}, each agent $i$'s estimate satisfies $y_s^i=\frac{1}{N}\sum_{j=1}^N r_{t_s}^j+\gamma_s$ with $|\gamma_s|\le 1/s$. Writing $r_{t_s}^j=x_{t_s}^\top\theta_*^j+\eta_{t_s}^j$ and defining $\zeta_{t_s}=\frac{1}{N}\sum_{j=1}^N\eta_{t_s}^j$, which is conditionally $R/\sqrt N$-sub-Gaussian, we have from \eqref{eq:rls}
\[
\hat{\theta}^{\mathrm{global},i}_s-\theta_*^{\mathrm{global}}
=\Sigma_s^{-1}\!\Big[\sum_{k=1}^s x_{t_k}\zeta_{t_k}
+\sum_{k=1}^s x_{t_k}\gamma_k-\lambda\theta_*^{\mathrm{global}}\Big].
\]
Standard RLS analysis with consensus errors $|\gamma_k|\le 1/k$ gives uniformly in $s$ and for each agent $i$
\[
\big\|\hat{\theta}^{\mathrm{global},i}_s-\theta_*^{\mathrm{global}}\big\|_{\Sigma_s}
\ \le\ \beta_s,
\]
where $\beta_s$ is from \eqref{eq:beta}. A union bound over $i\in[N]$ with $\delta_{\mathrm{conf}}=\delta/(2N)$ yields the claim.

\emph{(ii) Safe-set inclusion.}
Since $\theta_*^{\mathrm{global}}\in\mathcal{E}^i_{s-1}$ by part (i), for any $v\in\mathcal{E}^i_{s-1}$,
\begin{align*}
x^{*\top}v
&\ge x^{*\top}\hat{\theta}^{\mathrm{global},i}_{s-1}
  -\beta_{s-1}\|x^\star\|_{\Sigma_{s-1}^{-1}} \\
&\ge x^{*\top}\theta_*^{\mathrm{global}}
  -2\beta_{s-1}\|x^\star\|_{\Sigma_{s-1}^{-1}}.
\end{align*}
Using $\|x^\star\|_2\le L$ and $\|x^\star\|_{\Sigma_{s-1}^{-1}}\le \sqrt{L/\lambda_{\min}(\Sigma_{s-1})}$,
\[
x^{*\top}v
\ \ge\ r_{b,t_s}+\kappa_{b,t_s}
-2\beta_{s-1}\sqrt{L/\lambda_{\min}(\Sigma_{s-1})}.
\]
Under \eqref{eq:lambda-thresh}, the right-hand side is at least $r_{b,t_s}+\kappa_l-(\kappa_l+\alpha r_l)=(1-\alpha)r_{b,t_s}$. By \eqref{eq:safe_set}, this means $x^\star\in\mathcal{X}^{\mathrm{safe},i}_{s-1}$.

\textbf{Proof of Lemma~\ref{lem:num-cons}.}
Whenever a conservative action is played in episode $s$, either $\mathcal{X}^{\mathrm{safe},a(s)}_{s-1}=\emptyset$ or $\lambda_{\min}(\Sigma_{s-1})<(\tfrac{2L\,\beta_{s-1}}{\kappa_l+\alpha r_l})^2$. The randomized conservative action $x_t^{\text{cons}}=(1-\rho)x_{b,t}+\rho\zeta_t$ increases $\lambda_{\min}(\Sigma)$ in expectation. A matrix-Azuma inequality lower-bounds $\lambda_{\min}(\Sigma_s)$ in terms of the number of conservative episodes (similar to the proof of Theorem G.4 in \cite{moradipari2020stage}), and solving the resulting quadratic yields the explicit count in \eqref{eq:num-cons}.

\textbf{Proof of Theorem~\ref{thm:regret}.}
Let $\mathcal E_{\text{conf}}$ be the event in Theorem~\ref{thm:confidence}(i) (probability $\ge 1-\delta/2$) and $\mathcal E_{\text{count}}$ the event of Lemma~\ref{lem:num-cons} (probability $\ge 1-\delta/2$). We work on $\mathcal E=\mathcal E_{\text{conf}}\cap\mathcal E_{\text{count}}$.

\emph{Decomposition and communication accounting.}
The instantaneous per-episode regret $\Delta_s$ is at most
\[
\Delta_s\ \le\
\begin{cases}
2\beta_{s-1}\|x_{t_s}\|_{\Sigma_{s-1}^{-1}}, & \text{UCB episode},\\
\kappa_h+\rho(r_h+S), & \text{conservative episode}.
\end{cases}
\]
Each episode repeats its action for $1+q(s)$ rounds. Since $q(\cdot)$ is non-decreasing, $\sum_{s\le M}(1+q(s))\Delta_s\le (1+q(M))\sum_{s\le M}\Delta_s$.

\emph{UCB episodes.}
By Theorem~\ref{thm:confidence}(ii), $x^\star$ is feasible in every UCB episode. Summing $2\beta_{s-1}\|x_{t_s}\|_{\Sigma_{s-1}^{-1}}$ over UCB episodes and applying the elliptical potential lemma yields
\[
\sum_{s\in\mathcal N_M}
\bigl(x^{*\top}\theta_*^{\mathrm{global}}
      -x_{t_s}^\top\theta_*^{\mathrm{global}}\bigr)
\ \le\
2\beta_M\sqrt{\,2dM\log\!\big(1+\tfrac{ML^2}{\lambda d}\big)}.
\]

\emph{Conservative episodes.}
Each contributes at most $\kappa_h+\rho(r_h+S)$; the number is bounded by \eqref{eq:num-cons}. Combining the two parts and multiplying by $(1+q(M))$ gives \eqref{eq:regret-main}.

\section{Experiments}\label{sec:experiments}

\begin{figure*}[t]
\centering
\begin{minipage}{0.49\textwidth}
  \centering
  \includegraphics[width=\linewidth]{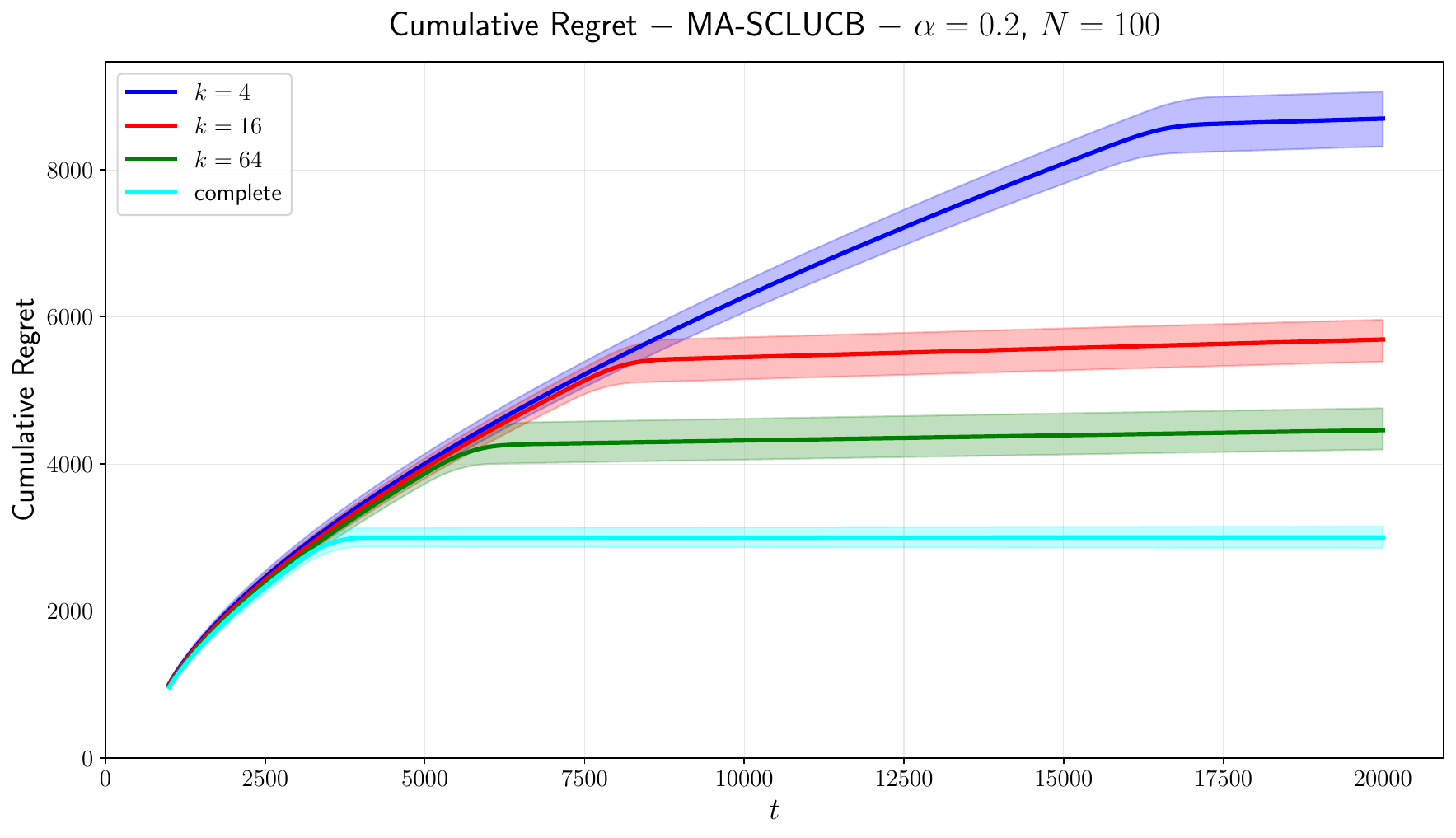}
  \subcaption{Cumulative regret vs.\ connectivity}
\end{minipage}
\hfill
\begin{minipage}{0.49\textwidth}
  \centering
  \includegraphics[width=\linewidth]{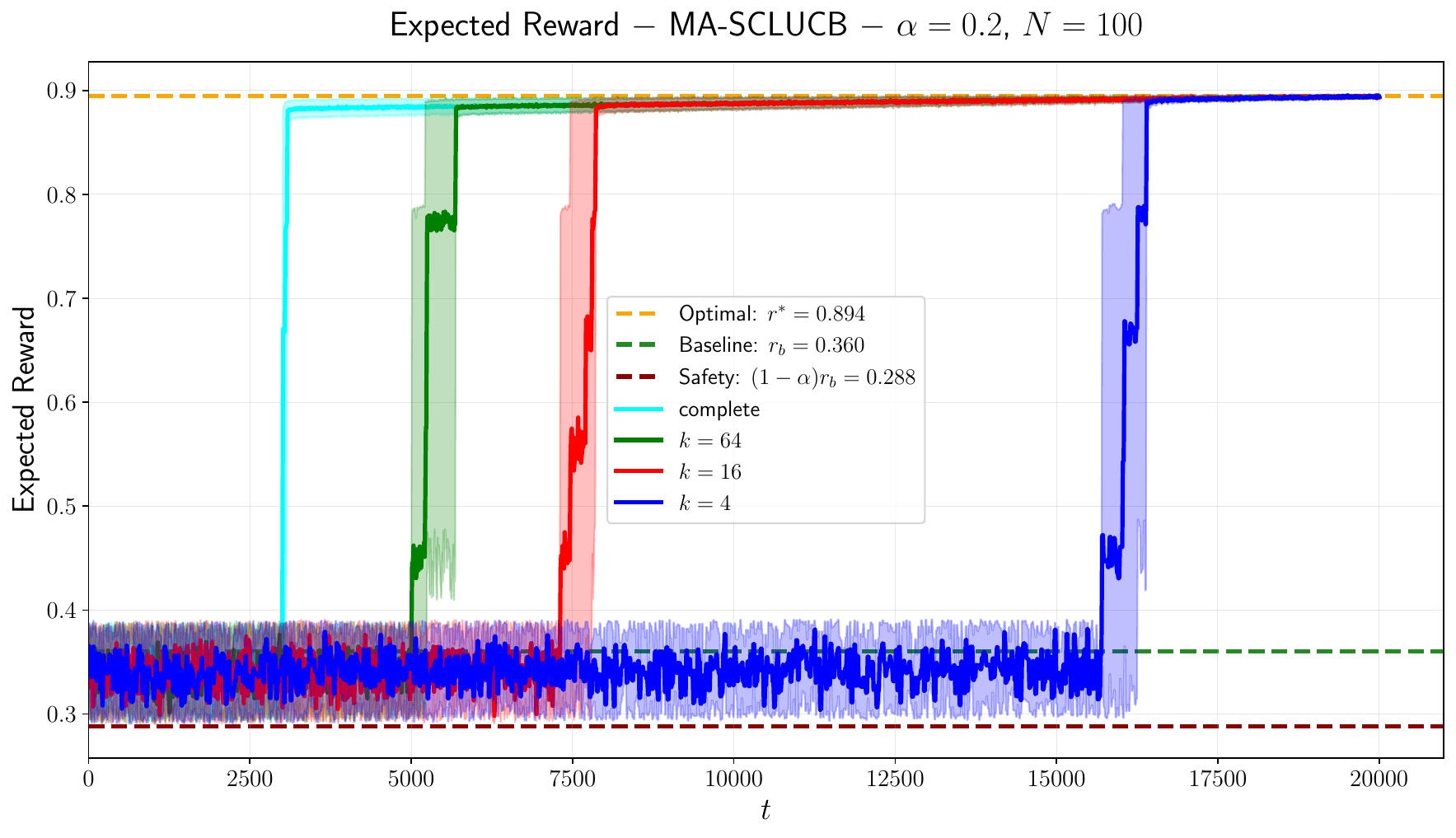}
  \subcaption{Expected reward and safety threshold}
\end{minipage}

\vspace{0.3cm}

\begin{minipage}{0.49\textwidth}
  \centering
  \includegraphics[width=\linewidth]{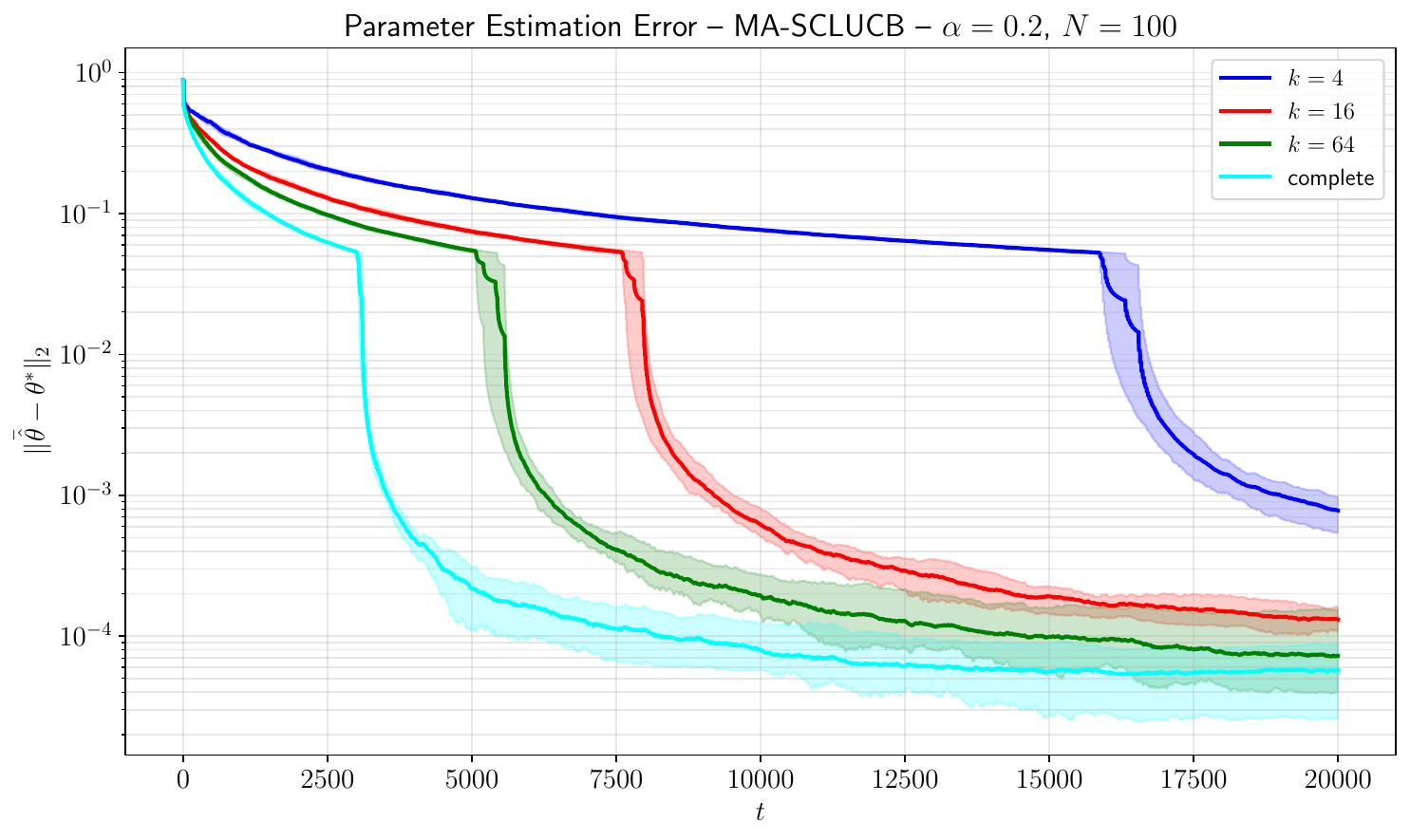}
  \subcaption{Parameter estimation convergence}
\end{minipage}
\hfill
\begin{minipage}{0.49\textwidth}
  \centering
  \includegraphics[width=\linewidth]{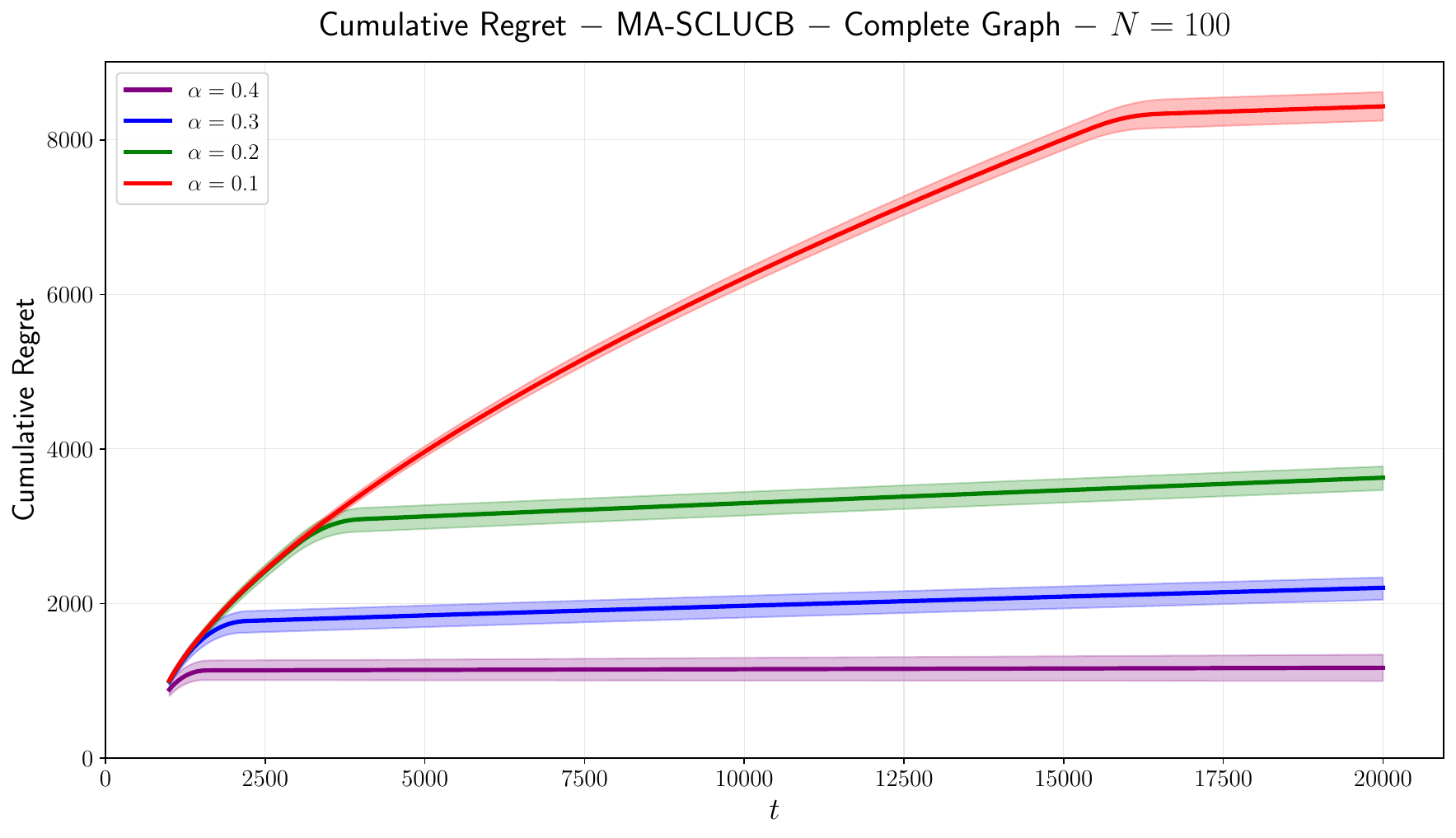}
  \subcaption{Cumulative regret vs.\ $\alpha$}
\end{minipage}

\caption{Experimental validation of MA-SCLUCB. (a-c) Impact of network connectivity on $k$-regular graphs with $N=100$, $\alpha=0.2$. (d) Effect of conservativeness parameter $\alpha$ on complete graphs with $N=100$.}
\label{fig:experiments}
\end{figure*}

We empirically validate MA-SCLUCB on synthetic linear bandit problems to demonstrate the theoretical insights from Section~\ref{sec:regret}. All experiments use $d=2$ dimensional problems with $T=20000$ rounds unless otherwise specified. We set hyperparameters $R=0.01$, $S=1.0$, $\lambda=0.1$, $\delta=0.01$, and $L=1.0$. Local reward parameters $\theta^i_*$ are sampled uniformly from the unit ball, and baseline actions are generated to ensure sub-optimality gaps satisfy Assumption~\ref{ass:baseline}. For action selection, we solve the convex optimization problem using the method in \cite{afsharrad2024convex}, which constructs the action set as a union of convex sets and applies $\ell_1$ relaxation for the safe set constraint—a standard approximation in conservative bandit algorithms \cite{moradipari2020stage, moradipari2019safe, cmacslb}. All results are averaged over 50 independent runs.

\subsection{Impact of Network Connectivity}

We first investigate how graph structure affects performance. Figure~\ref{fig:experiments}(a-c) shows results for $k$-regular graphs with $N=100$ agents and $\alpha=0.2$, varying $k\in\{4, 16, 64, 99\}$ (where $k=99$ is the complete graph). Figure~\ref{fig:experiments}(a) demonstrates that cumulative regret decreases with connectivity: better-connected networks (larger $k$, smaller $|\lambda_2|$) achieve faster convergence to the optimal policy. This validates our theoretical bound's dependence on $|\lambda_2|$ through the communication overhead term.

Figure~\ref{fig:experiments}(b) verifies safety compliance across all network configurations. The expected reward (solid lines) consistently remains above the conservative threshold $(1-\alpha)r_{b,t}$ (dashed line), confirming that MA-SCLUCB never violates the stage-wise constraint. The plots also reveal the algorithm's two-phase behavior: an initial \emph{exploration phase} where conservative actions dominate to build confidence regions, followed by an \emph{exploitation phase} where UCB actions are selected once sufficient exploration has occurred (corresponding to the condition in Theorem~\ref{thm:confidence}(ii)).

Figure~\ref{fig:experiments}(c) shows parameter estimation error $\|\frac{1}{N}\sum_{i=1}^N \hat{\theta}^{\text{global},i}_s - \theta_*^{\text{global}}\|_2$ over time. All configurations converge to the true global parameter, but convergence accelerates with connectivity, as predicted by our analysis: well-connected networks require fewer communication rounds per episode to achieve accurate consensus.

\subsection{Effect of Conservativeness Level}

Figure~\ref{fig:experiments}(d) illustrates how the conservativeness parameter $\alpha$ affects learning. We test $\alpha\in\{0.1, 0.2, 0.3, 0.4\}$ on complete graphs with $N=100$. Smaller $\alpha$ values impose stricter safety requirements, forcing the algorithm to play conservative actions longer before sufficient exploration permits UCB action selection. This translates directly to higher cumulative regret and slower convergence. The trend aligns with our regret analysis: tighter constraints increase the threshold $k_t$ in \eqref{eq:lambda-thresh}, delaying the transition from conservative to UCB episodes.

\subsection{Network Size Scaling}

To verify the $1/\sqrt{N}$ improvement predicted by Theorem~\ref{thm:regret}, we examine parameter estimation accuracy as a function of network size. Table~\ref{tab:network_size} reports $\|\frac{1}{N}\sum_{i=1}^N \hat{\theta}^{\text{global},i}_{1000} - \theta_*^{\text{global}}\|_2$ after $T=1000$ rounds for complete graphs with $N\in\{1, 10, 100, 1000\}$. Estimation error decreases substantially as $N$ grows, confirming that distributed collaboration—despite local communication constraints—yields meaningful statistical gains from averaging across agents' observations.

\begin{table}[t]
\centering
\caption{Parameter estimation error after 1000 rounds vs.\ network size on complete graphs with $\alpha=0.2$.}
\label{tab:network_size}
\begin{tabular}{c|cccc}
\hline
$N$ & 1 & 10 & 100 & 1000 \\
\hline
Error & 0.00326 & 0.00106 & 0.000554 & 0.000517 \\
\hline
\end{tabular}
\end{table}

\section{Conclusion}\label{sec:conclusion}

We studied the multi-agent stage-wise conservative linear bandit problem, where networked agents must collaboratively learn while maintaining safety guarantees at every round. We proposed MA-SCLUCB, an episodic algorithm that alternates between action selection and consensus-building phases, and proved a regret bound of $\tilde{O}\left(\frac{d}{\sqrt{N}}\sqrt{T}\cdot\frac{\log(NT)}{\sqrt{\log(1/|\lambda_2|)}}\right)$. Our analysis reveals three key insights: (i) distributed collaboration yields a fundamental $\frac{1}{\sqrt{N}}$ statistical advantage despite local communication constraints, (ii) the communication overhead grows only logarithmically for well-connected networks, and (iii) stage-wise safety constraints add only lower-order regret terms. Experimental results validate these theoretical findings across varying network structures, conservativeness levels, and network sizes. This work demonstrates that safety-constrained distributed learning can achieve near-optimal performance in reasonably connected networks, opening avenues for safe collaborative decision-making in applications such as recommendation systems and autonomous systems. Future work could explore heterogeneous communication costs, time-varying network topologies, or extensions to nonlinear reward models.

\bibliographystyle{IEEEtran}
\bibliography{refs.bib}

@inproceedings{moradipari2020stage,
  title={Stage-wise conservative linear bandits},
  author={Moradipari, Ahmadreza and Thrampoulidis, Christos and Alizadeh, Mahnoosh},
  booktitle={Advances in Neural Information Processing Systems},
  volume={33},
  pages={19487--19498},
  year={2020}
}

@INPROCEEDINGS{cmacslb,
  author={Afsharrad, Amirhossein and Oftadeh, Parisa and Moradipari, Ahmadreza and Lall, Sanjay},
  booktitle={2025 American Control Conference (ACC)}, 
  title={Cooperative Multi-Agent Constrained Stochastic Linear Bandits}, 
  year={2025},
  volume={},
  number={},
  pages={3614-3621},
  doi={10.23919/ACC63710.2025.11107780}
}

@article{bubeck2012regret,
  title={Regret analysis of stochastic and nonstochastic multi-armed bandit problems},
  author={Bubeck, S{\'e}bastien and Cesa-Bianchi, Nicolo},
  journal={arXiv preprint arXiv:1204.5721},
  year={2012}
}

@article{khezeli2019safe,
  title={Safe Linear Stochastic Bandits},
  author={Khezeli, Kia and Bitar, Eilyan},
  journal={arXiv preprint arXiv:1911.09501},
  year={2019}
}

@inproceedings{abbasi2011improved,
  title={Improved algorithms for linear stochastic bandits},
  author={Abbasi-Yadkori, Yasin and P{\'a}l, D{\'a}vid and Szepesv{\'a}ri, Csaba},
  booktitle={Advances in Neural Information Processing Systems},
  pages={2312--2320},
  year={2011}
}

@inproceedings{filippi2010parametric,
  title={Parametric bandits: The generalized linear case},
  author={Filippi, Sarah and Cappe, Olivier and Garivier, Aur{\'e}lien and Szepesv{\'a}ri, Csaba},
  booktitle={Advances in Neural Information Processing Systems},
  pages={586--594},
  year={2010}
}

@inproceedings{li2017provably,
  title={Provably optimal algorithms for generalized linear contextual bandits},
  author={Li, Lihong and Lu, Yu and Zhou, Dengyong},
  booktitle={Proceedings of the 34th International Conference on Machine Learning-Volume 70},
  pages={2071--2080},
  year={2017},
  organization={JMLR. org}
}

@article{Tsitsiklis,
author = {Rusmevichientong, Paat and Tsitsiklis, John N.},
title = {Linearly Parameterized Bandits},
journal = {Mathematics of Operations Research},
volume = {35},
number = {2},
pages = {395-411},
year = {2010},
doi = {10.1287/moor.1100.0446},
}

@inproceedings{dani2008stochastic,
  title={Stochastic linear optimization under bandit feedback},
  author={Dani, Varsha and Hayes, Thomas P and Kakade, Sham M},
  booktitle={21st Annual Conference on Learning Theory},
  number={101},
  pages={355--366},
  year={2008}
}

@article{Auer,
 author = {Auer, Peter and Cesa-Bianchi, Nicol\`{o} and Fischer, Paul},
 title = {Finite-time Analysis of the Multiarmed Bandit Problem},
 journal = {Mach. Learn.},
 issue_date = {May-June 2002},
 volume = {47},
 number = {2-3},
 month = may,
 year = {2002},
 issn = {0885-6125},
 pages = {235--256},
 numpages = {22},
 url = {https://doi.org/10.1023/A:1013689704352},
 doi = {10.1023/A:1013689704352},
 acmid = {599677},
 publisher = {Kluwer Academic Publishers},
 address = {Hingham, MA, USA},
}

@incollection{vanroy,
title = {Conservative Contextual Linear Bandits},
author = {Kazerouni, Abbas and Ghavamzadeh, Mohammad and Abbasi, Yasin and Van Roy, Benjamin},
booktitle = {Advances in Neural Information Processing Systems 30},
editor = {I. Guyon and U. V. Luxburg and S. Bengio and H. Wallach and R. Fergus and S. Vishwanathan and R. Garnett},
pages = {3910--3919},
year = {2017},
publisher = {Curran Associates, Inc.},
url = {http://papers.nips.cc/paper/6980-conservative-contextual-linear-bandits.pdf}
}

@article{abeille2017linear,
  title={Linear Thompson sampling revisited},
  author={Abeille, Marc and Lazaric, Alessandro and others},
  journal={Electronic Journal of Statistics},
  volume={11},
  number={2},
  pages={5165--5197},
  year={2017},
  publisher={The Institute of Mathematical Statistics and the Bernoulli Society}
}

@inproceedings{kaufmann2012thompson,
  title={Thompson sampling: An asymptotically optimal finite-time analysis},
  author={Kaufmann, Emilie and Korda, Nathaniel and Munos, R{\'e}mi},
  booktitle={International Conference on Algorithmic Learning Theory},
  pages={199--213},
  year={2012},
  organization={Springer}
}

@article{thompson1933likelihood,
  title={On the likelihood that one unknown probability exceeds another in view of the evidence of two samples},
  author={Thompson, William R},
  journal={Biometrika},
  volume={25},
  number={3/4},
  pages={285--294},
  year={1933},
  publisher={JSTOR}
}

@article{russo2016information,
  title={An information-theoretic analysis of thompson sampling},
  author={Russo, Daniel and Van Roy, Benjamin},
  journal={The Journal of Machine Learning Research},
  volume={17},
  number={1},
  pages={2442--2471},
  year={2016},
  publisher={JMLR. org}
}

@inproceedings{bubeck2016multi,
  title={Multi-scale exploration of convex functions and bandit convex optimization},
  author={Bubeck, S{\'e}bastien and Eldan, Ronen},
  booktitle={Conference on Learning Theory},
  pages={583--589},
  year={2016}
}

@inproceedings{agrawal2013thompson,
  title={Thompson sampling for contextual bandits with linear payoffs},
  author={Agrawal, Shipra and Goyal, Navin},
  booktitle={International Conference on Machine Learning},
  pages={127--135},
  year={2013}
}

@inproceedings{moradipari2018learning,
  title={Learning to dynamically price electricity demand based on multi-armed bandits},
  author={Moradipari, Ahmadreza and Silva, Cody and Alizadeh, Mahnoosh},
  booktitle={2018 IEEE Global Conference on Signal and Information Processing (GlobalSIP)},
  pages={917--921},
  year={2018},
  organization={IEEE}
}

@inproceedings{wu2016conservative,
  title={Conservative bandits},
  author={Wu, Yifan and Shariff, Roshan and Lattimore, Tor and Szepesv{\'a}ri, Csaba},
  booktitle={International Conference on Machine Learning},
  pages={1254--1262},
  year={2016}
}

@article{moradipari2019safe,
  title={Safe Linear Thompson Sampling with Side Information},
  author={Moradipari, Ahmadreza and Amani, Sanae and Alizadeh, Mahnoosh and Thrampoulidis, Christos},
  journal={arXiv},
  pages={arXiv--1911},
  year={2019}
}

@article{DBLP:journals/corr/LandgrenSL16,
  author       = {Peter Landgren and
                  Vaibhav Srivastava and
                  Naomi Ehrich Leonard},
  title        = {Distributed Cooperative Decision-Making in Multiarmed Bandits: Frequentist
                  and Bayesian Algorithms},
  journal      = {CoRR},
  volume       = {abs/1606.00911},
  year         = {2016},
  url          = {http://arxiv.org/abs/1606.00911},
  eprinttype    = {arXiv},
  eprint       = {1606.00911},
  bibsource    = {dblp computer science bibliography, https://dblp.org}
}

@INPROCEEDINGS{9143736,
  author={Madhushani, Udari and Leonard, Naomi Ehrich},
  booktitle={2020 European Control Conference (ECC)}, 
  title={A Dynamic Observation Strategy for Multi-agent Multi-armed Bandit Problem}, 
  year={2020},
  volume={},
  number={},
  pages={1677-1682},
  doi={10.23919/ECC51009.2020.9143736}}

@inproceedings{ijcai2017p24,
  author    = {Mithun Chakraborty and Kai Yee Phoebe Chua and Sanmay Das and Brendan Juba},
  title     = {Coordinated Versus Decentralized Exploration In Multi-Agent Multi-Armed Bandits},
  booktitle = {Proceedings of the Twenty-Sixth International Joint Conference on
               Artificial Intelligence, {IJCAI-17}},
  pages     = {164--170},
  year      = {2017},
  doi       = {10.24963/ijcai.2017/24},
  url       = {https://doi.org/10.24963/ijcai.2017/24},
}

@misc{landgren2020distributed,
      title={Distributed Cooperative Decision Making in Multi-agent Multi-armed Bandits}, 
      author={Peter Landgren and Vaibhav Srivastava and Naomi Ehrich Leonard},
      year={2020},
      eprint={2003.01312},
      archivePrefix={arXiv},
      primaryClass={math.OC}
}

@ARTICLE{5738217,
  author={Anandkumar, Animashree and Michael, Nithin and Tang, Ao Kevin and Swami, Ananthram},
  journal={IEEE Journal on Selected Areas in Communications}, 
  title={Distributed Algorithms for Learning and Cognitive Medium Access with Logarithmic Regret}, 
  year={2011},
  volume={29},
  number={4},
  pages={731-745},
  doi={10.1109/JSAC.2011.110406}}

@article{Kar2011BanditPI,
  title={Bandit problems in networks: Asymptotically efficient distributed allocation rules},
  author={Soummya Kar and H. Vincent Poor and Shuguang Cui},
  journal={IEEE Conference on Decision and Control and European Control Conference},
  year={2011},
  pages={1771-1778},
  url={https://api.semanticscholar.org/CorpusID:1544223}
}

@inproceedings{afsharrad2024convex,
  title={Convex methods for constrained linear bandits},
  author={Afsharrad, Amirhossein and Moradipari, Ahmadreza and Lall, Sanjay},
  booktitle={2024 European Control Conference (ECC)},
  pages={2111--2118},
  year={2024},
  organization={IEEE}
}

\end{document}